\documentclass[conference]{IEEEtran}

\usepackage{amsfonts,amsmath,amsthm}
\usepackage{graphicx,xspace,epsfig,syntonly,dsfont}
\usepackage{url,cite,bm,bbm}
\usepackage{algorithmicx}
\usepackage{balance}
\usepackage{subfigure}
\usepackage{stfloats}
\usepackage{diagbox}

\long\def\symbolfootnote[#1]#2{\begingroup
\def\thefootnote{\fnsymbol{footnote}}
\footnote[#1]{#2}\endgroup}

\IEEEoverridecommandlockouts

\allowdisplaybreaks[4]

\title{\Large \bf  
Mitigating Greenhouse Gas Emissions Through  \\ Generative Adversarial Networks Based Wildfire Prediction}
\author{
    \IEEEauthorblockN{Sifat Chowdhury\IEEEauthorrefmark{1}, Kai Zhu\IEEEauthorrefmark{2} and Yu Zhang\IEEEauthorrefmark{3}}
     \IEEEauthorblockA{\IEEEauthorrefmark{1}Dept. of Electrical and Computer Engineering, University of California, Santa Cruz,\, \texttt{schowdh6@ucsc.edu}}
    \IEEEauthorblockA{\IEEEauthorrefmark{2}Dept. of Environmental Studies, University of California, Santa Cruz,\,  \texttt{kai.zhu@ucsc.edu}}
     \IEEEauthorblockA{\IEEEauthorrefmark{3}Dept. of Electrical and Computer Engineering, University of California, Santa Cruz,\, \texttt{zhangy@ucsc.edu}}

    \thanks{This work was supported in part by the Faculty Research Grant of UC Santa Cruz, 2019 Seed Fund Award from CITRIS and the Banatao Institute at the University of California, and the Hellman Fellowship.}
}

\begin{document}
\maketitle

\begin{abstract}

Over the past decade, the number of wildfire has increased significantly around the world, especially in the State of California. The high-level concentration of greenhouse gas (GHG) emitted by wildfires aggravates global warming that further increases the risk of more fires. Therefore, an accurate prediction of wildfire occurrence greatly helps in preventing large-scale and long-lasting wildfires and reducing the consequent GHG emissions. Various methods have been explored for wildfire risk prediction. However, the complex correlations among a lot of natural and human factors and wildfire ignition make the prediction task very challenging. In this paper, we develop a deep learning based data augmentation approach for wildfire risk prediction. We build a dataset consisting of diverse features responsible for fire ignition and utilize a conditional tabular generative adversarial network to explore the underlying patterns between the target value of risk levels and all involved features. For fair and comprehensive comparisons, we compare our proposed scheme with five other baseline methods where the former outperformed most of them. To corroborate the robustness, we have also tested the performance of our method with another dataset that also resulted in better efficiency. By adopting the proposed method, we can take preventive strategies of wildfire mitigation to reduce global GHG emissions.
\end{abstract}

\section{Introduction}\label{sec:intro}

As one of the most destructive events for both human and nature, wildland fire is a major source responsible for GHG emission. It poses a great threat to many aspects of our life, ranging from poor air quality and loss of habitation to the damage of lots of valuable assets. The frequency of wildfires has increased by a factor of four compared to 1970 which largely owes to climate change. It is estimated that annually between 350 and 450 million hectares of forest and grassland are burnt by wildfires \cite{WF_emission}. It is evident that global warming contributes to the increasing number of wildfires, which turn the forests as carbon sources instead of being carbon sinks. 
In general, wildfires emit 5 to 30 tonnes of carbon per hectare; therefore, making the annual carbon production between 1.75 and 13.5 billion metric tonnes. Zooming into California, one of the worst sufferers from wildfires, got 4.2 million acres of land burnt in 2020 alone (which is roughly 4\% of the total area of the state) \cite{CA_4}. According to Global Fire Emission Database (GFED), California wildfires had generated more than 91 million metric tons of GHG, which is 25\% more than annual emissions from fossil fuels in the state. 

\begin{figure}[t!]
\centering
  \includegraphics[scale=0.20]{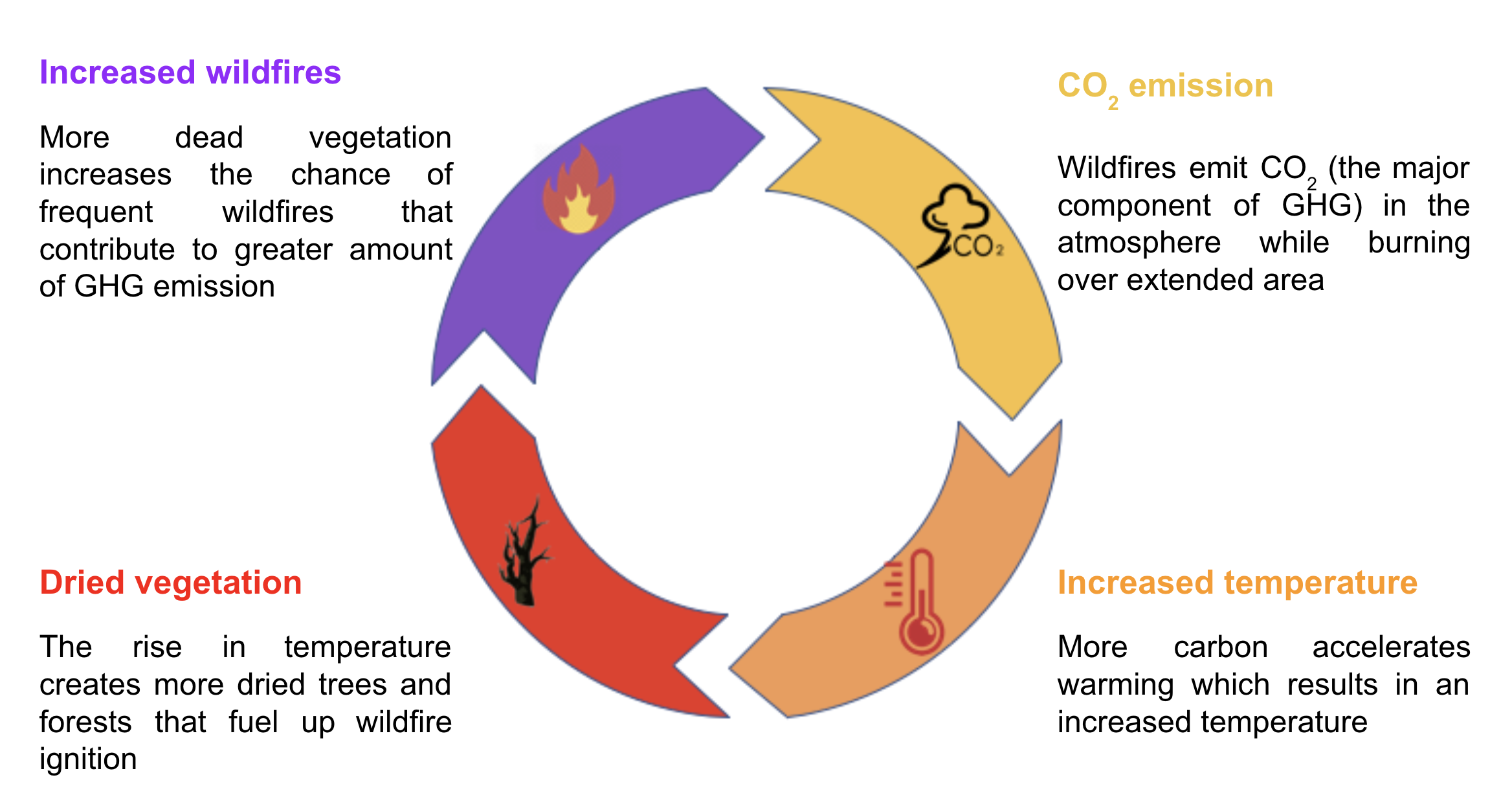}
  \caption{Loop between fire events and GHG emission.}
  \label{fig:cycle}
  
\end{figure}

Forests capture carbon from the atmosphere in the soils and trees. When they burn, a large amount of carbon gets released. This GHG emission from wildfires forms a vicious cycle represented in Figure \ref{fig:cycle}. More emission results in more change in the climate and increases the temperature. A rise in temperature makes the vegetation dryer, more flammable, susceptible for burning for longer period and lessening the growth of newer vegetation, thus creating a feedback loop of increased GHG emission. 

In order to curb climate change, we need to reduce the incidence of wildfires that shoots up the amount of GHG in the environment. For doing so, we must be able to accurately predict the occurrences of fire events. This is the main objective of this paper. 

Researchers from different disciplines have explored various ways to forecast fire locations. Preisler et al. produced a map of potential fire distribution of the whole USA using gridded satellite data and surface observation   \cite{Preisler_2009}. Later they performed a spatially explicit forecast for modelling the the probability of a large wildland fire \cite{Preisler_2011}. Both of the papers have used remote sensing data to make the prediction in conventional approach. Gasull et al. utilized wireless sensor networks to predict the risk of a wildfire but this is not an effective way in terms of both cost and precision \cite{wireless}. Machine learning (ML) techniques have also been used in prediction tasks. One of the recent works by Nadeem et al. used a Lasso-logistic framework for human and lightning caused forest fire occurrences \cite{nadeem}. Sayad et al. have shared a new dataset consisting of some of the underlying features causing wildfire and used some baseline ML algorithms for predicting fire locations \cite{Sayad2019}. Another method for training deep neural networks using a validation-loss (VL) weight selection strategy on imbalanced datasets have been used by Langford et al. to detect wildfire events \cite{dnn_Langford}. A physics-assisted recurrent neural network model for mapping fuel moisture content was introduced by Rao et al. to characterize wildfire risk \cite{rnn_Rao}
None of these works has considered all the important aspects together in a holistic manner that are responsible for initiating a fire. This relationship is indeed very complex as there are a lot of features and they are interdependent on each other. Unraveling this complex relation is the key to a successful prediction of fire occurrences.

Deep learning is gaining much popularity in identifying complex patterns due to its supremacy of accuracy when trained with a massive amount of data. In this paper, we have investigated the underlying relation between a wildland fire and its corresponding factors to predict the fire locations one day ahead. We have also checked the performance of our proposed method on another dataset used by \cite{Sayad2019} in order to prove the efficacy of our approach. The major contributions of our paper are outlined as follows:
\begin{itemize}
  \item We construct a dataset that consists of the relevant features (meteorological factors, vegetation status, topology and proximity to nearby power lines) combined together to correctly identify the causation of a wildfire in a comprehensive way. 
  \item We utilize a state-of-the-art deep learning technique, generative adversarial network (GAN) for conditional generation of tabular data to generate sufficient inputs from the original dataset for properly train ML models. To the best of our knowledge, no one has implemented a GAN based technique in a dataset that has a wide variety of fire risk elevating factors that gives a more accurate prediction of wildfire occurrences.
\end{itemize}

The remainder of this paper is organized as follows. Section II provides the details of our dataset construction. In Section III, the architecture of the conditional tabular GAN (CTGAN) is discussed. Simulation studies are presented in Section IV. Finally, Section V discusses how this prediction can be utilized in reducing carbon footprints along with the future work.\\

\section{Data Set Construction} \label{sec:DS formation}

\subsubsection{Set of features} \label{ds:feature}

For predicting fire locations, we first need to analyze the complex dynamics between different contributing factors and how they lead to a fire ignition. Discovering that part is not an easy task as there are a lot of reasons by which a fire can be initiated. So, in order to model the relationship, we have decided to build a dataset from scratch consisting of a wide variety of features. Among them, weather or meteorological factors have a direct, strong causation. Also, it is evident from the fact that over the past few decades, climate change has given rise to the frequency of all sizes of fires, large, small and big, all over the world. We have collected the data for temperature, precipitation, surface pressure, direction and speed of wind and humidity of a specific geographical location. Besides that, naturally occurring wildfires can easily be sparked by dry weather and droughts since dry vegetation acts as a flammable fuel while warm temperature encourages combustion. Therefore, vegetation indices play a prominent role in determining fire hotspots. We have gathered the data for 2 such indices, which are normalized difference vegetation index (NDVI) and enhanced vegetation index (EVI). Both weather and vegetation data can be acquired from weather stations placed in certain locations. However, a big issue of station based data is that the value of a particular weather variable is assumed constant over a wide range of areas under the coverage of that station. Hence, precise data may not be available because we do not have so many stations for reasonably spaced locations. Installing lots of sensors can be a solution that is too expensive for many remote rural areas. By contrast, utilizing satellite images facilitate this purpose by maintaining a good trade-off between precision and cost. 

Another important component that influences the fire behaviour of a region is the topography. It is the physical feature of a place which is generally static (unless changed by men or some natural  disasters like hurricane, tornado etc.) and is opposite of weather which is ever changing. However, it interacts with wind speed and direction interestingly and thus eventually impacts fire initiation. In the slopes, the wind usually blow up during the day and less dense air (reflected by the surface) rises up along the slope. Hence, the steeper the slope, the quicker the hot air will flow uphill and preheat the flammable fuel to its ignition temperature. Thus the uphill zones are more prone to fire ignition during the daytime. The scenario is opposite at night when the cooler winds start to flow downhill. For this reason, we have incorporated the topographic information of the region in our dataset.

Last but not the least, one of the common causes of wildfire ignition is electrical equipment and power line failures. High temperature arcing created during a high impedance fault can ignite a proximate vegetation and other combustible materials. Other than that during high wind, trees and branches may fall across power lines and can result in two conductors coming in contact with each other. It produces high energy arcing and ejects hot metal particles that can eventually lead to a wildfire ignition. Considering this, we have included the distance between fire locations and the nearest power lines in our dataset to account for the chances of fire ignition due to power system equipment failure. 

\begin{figure}[t!]
\centering
  \includegraphics[scale=0.20]{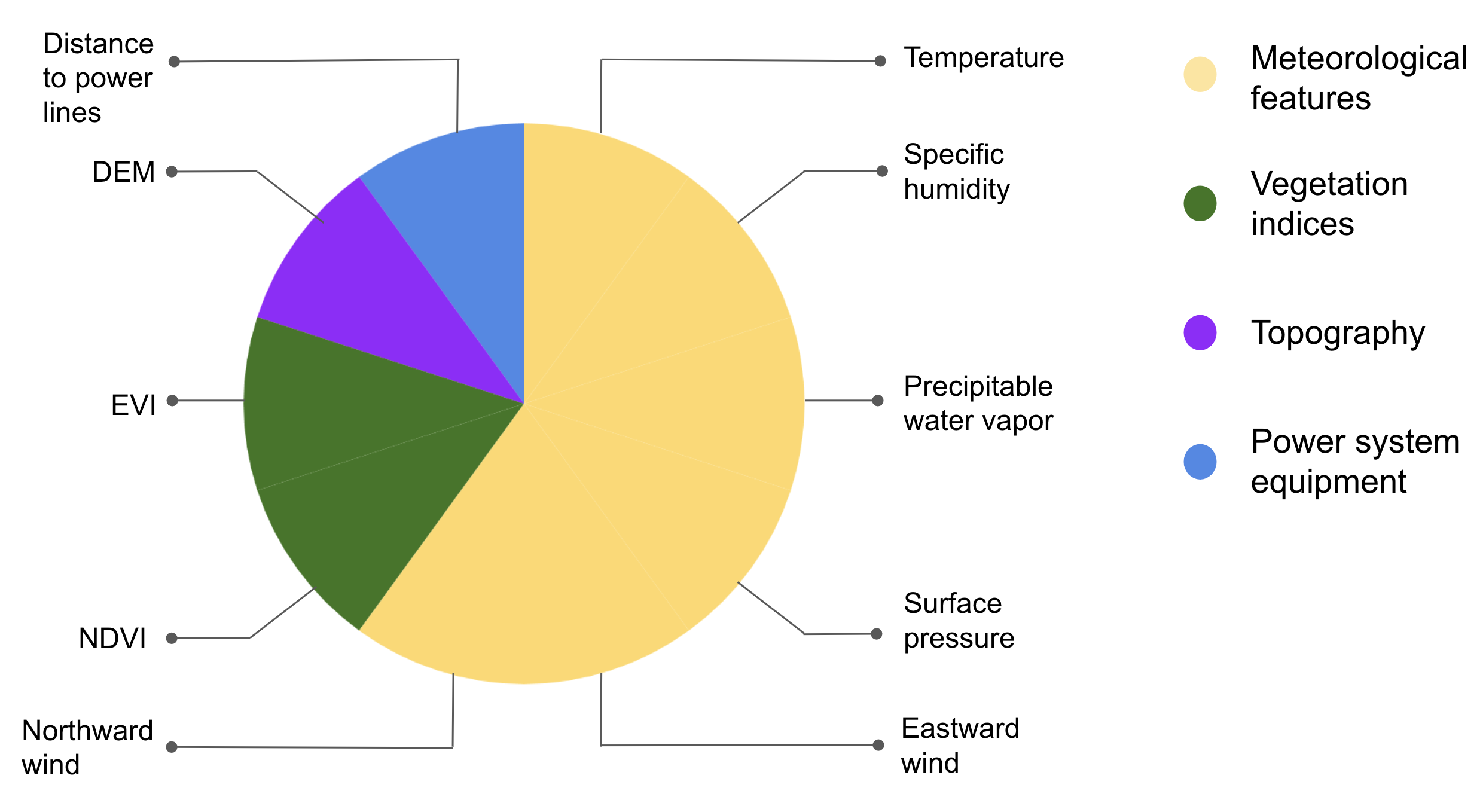}
  \caption{Features used in our dataset. (DEM= digital elevetaion model, NDVI= normalized difference vegetation index, EVI= enhanced vegetation index).}
  \label{fig:features}
  \end{figure}

The complete set of features used in our constructed dataset is shown in Figure \ref{fig:features}.\\

\subsubsection{Data collection} \label{ds:col}
Two most destructive wildfires in CA, `Camp' and `Tubbs', respectively occurred in Butte county, in November 2018 and  in Napa, Sonoma county in October, 2017 were used as the test case for our dataset. To start with, we collected the location of the fire points (latitude-longitude) from the earth observation data by NASA that used MODIS- Terra (MOD14) product to locate the fire pixel \cite{fire_data}. After that we projected those geographic points in QGIS and added some other points those are not on fire. In the dataset, we labeled these points as `No fire' while the points derived from \cite{fire_data} are labeled as `Fire'. In total, there are 849 data points where 420 data points hold the `Fire' label while the rest 429 points are labeled as `No fire'. As we are making a day ahead forecast, satellite images for the previous day containing the weather parameters i.e. temperature, specific humidity, precipitable water vapor, surface pressure, eastward and north wind, were collected from MERRA-2 (Modern-Era Retrospective analysis for Research and Applications version 2) for both the `Fire' and `No fire' points \cite{weather}. The vegetation indices for the same locations were collected from USGS that used Landsat 8 satellite, and having a 30m spatial resolution \cite{vegetation}. Digital elevation model (DEM) was used to get the topographic data of the points from USGS as well \cite{topography}. Finally, the information regarding the location of electric transmission lines were acquired from California Energy Commission \cite{power_line_data}.\\

\subsubsection{Data processing} \label{ds:proc}
The next step was to process the satellite images (in .tif format) and extract the pixel values to get different weather features (the same applies for vegetation indices and DEM). We used the ‘rgdal’ library of RStudio for image processing and value extraction. The Landsat-8 images for vegetation has 9 spectral bands among which band 2 (Visible Blue), band 4 (Visible Red) and band 5 (Near InfraRed or NIR) was used to extract NDVI and EVI.
The NDVI and EVI are calculated from these individual measurements according to the following equation:
\begin{align}
\textrm{NDVI} &= \frac{\textrm{NIR}-\textrm{Red}}{\textrm{NIR}+\textrm{Red}}   \\
\textrm{EVI} &= \frac{2.5 \times \textrm{NIR-Red}} {\textrm{NIR}+6 \times \textrm{Red}-7.5 \times \textrm{Blue}+1}.
\end{align}

\section{Prediction Using CTGAN } \label{ctgan}
Traditional ML approaches such as naive Bayes, decision trees (DF), support vector machines (SVM) often provide simple solutions to various classification tasks. However, disentangling the complex dynamics between wildfire occurrence and its underlying factors is a challenging task, which requires better techniques along with more data. Deep learning (DL) is such a technique where the `learning' part happens in the hidden layers of a neural network and `deep' refers to the number of those layers. As the `depth' of the network increases, the more insights about the prediction analysis can be found. One of the biggest hurdles towards utilizing these models is the limited number of data that need to feed into the network. 
We use remote sensing data for wildfire prediction. In this case, there is no simple way to extract more data from a high resolution source for a given location within a specific time. Hence, in order to enlarge the dataset, we need to generate synthetic data from the original set via an appropriate data augmentation technique. 

Generative adversarial network (GAN) is a DL based supervised learning approach, which discovers and learns the pattern in a dataset by its own and produces new data samples. Typically, a basic GAN model consists of two sub-networks: generator and discriminator. The generator generates new samples while the discriminator aims at distinguishing fake data points from real ones. Training a GAN is a minimax game:
\begin{align}
    \min_G\max_D~V(D,G) = &\mathbb{E}_{\mathbf{x}\sim p_{\mathrm{data}(\mathbf{x})}}[\log D(\mathbf{x})] + \notag  \\ &\mathbb{E}_{\mathbf{z}\sim p_{\mathbf{z}}(\mathbf{z})}[\log (1-D(G(\mathbf{z})))],
\end{align}
where the first term is entropy that the data from real distribution $p_{\mathrm{data}(\mathbf{x})}$ passes through the discriminator $D$. The second term is entropy that the data from random input $p_{\mathbf{z}}(\mathbf{z})$ passes through the generator $G$ that generates a fake sample, which is then passed through the discriminator $D$ to identify the genuineness. Overall, the discriminator aim to maximize the function $V$ while the generator tries to minimize it.
Essentially, the Jensen-Shannon divergence (or many other possible entropy metrics) between generator and data distributions is minimized. 
The inherent competition between the generator and the discriminator drives both to improve, until fake data are indistinguishable from genuine ones. As a side effect, the generator can draw high-quality realistic samples.

There are  different types of GANs used for a wide variety of applications. Chen et al. proposed InfoGAN - an information-theoretic extension to the GAN that is able to learn disentangled representations in an unsupervised manner \cite{infogan}. WGAN is another variant that uses Wassertain distance in the loss function \cite{wgan}. Another important adaptation is conditional generative adversarial networks (CGAN) that  uses an extra label information as an input to condition on both the generator and discriminator and can generate new samples conditioned on class labels \cite{cgan}. 

Although GAN offers a great flexibility in modeling distributions of image data, there are certain challenges when it comes to model tabular format of data containing a mixture of discrete and continuous columns. First, pixel values of images usually follow a Gaussian distribution which can be normalized and modeled by using a min-max transformation. However continuous values in tabular data do not follow such a distribution and the min-max transformation can lead to the issue of vanishing gradient. Second, the continuous columns may have multimodal distributions. It has been shown that the vanilla version of GAN cannot model all modes on a simple tabular dataset \cite{multimodal}. Hence, modeling continuous columns is a challenging task.

\subsection {Conditional tabular GAN}
To overcome the aforementioned challenges, Xu et al. proposed a conditional tabular GAN (CTGAN) model, which uses mode-specific normalization to combat the non-Gaussian and multimodal distributions of continuous columns and a conditional generator to accurately model the discrete columns in a tabular dataset. The following steps are required to implement the mode-specific normalization \cite{ctgan}. 
\begin{enumerate}
    \item For each column of the dataset, apply a variational Gaussian mixture model (VGM) to estimate the number of modes $K_i$ and fit a Gaussian mixture. 
    Let $c_{j,i}$ denote the $j$-th element of column $C_{i}$. Then, the learned Gaussian mixture model is given as 
    \begin{equation}
        \mathbb{P}_{C_{i}}(c_{j,i}) = \sum_{k=1}^{K_i} \mu_{k} \mathcal{N} {(c_{j,i};\eta_{k},\phi_{k})},
    \end{equation}
    where $\mu_{k}$ and $\phi_{k}$ are the weight and standard deviation of the mode $\eta_{k}$ for $k=1,2,\ldots,K_i$.
    \item For each value of $c_{j,i}$, calculate the probability of it coming from each mode. The probability densities are computed as 
    \begin{equation}
        \rho_{k} = \mu_{k} \mathcal{N} {(c_{j,i};\eta_{k},\phi_{k})},\, k=1,2,\ldots,K_i.
    \end{equation}
     \item Based on the probability densities, sample a mode to normalize the value. Specifically, $c_{j,i}$ is represented by a one-hot encoded vector $\beta_{j,i}$ representing the sampled mode and a scalar $\alpha_{j,i}$ = $\frac{c_{j,i}-\eta_{k}}{4\phi_{k}}$ to represent the value within the $k$-th picked mode.
\end{enumerate}
    
    In general, any row $r_j$ can be represented as 
    \begin{equation}
    r_{j} = \alpha_{j,1} \oplus \beta_{j,1} \oplus \alpha_{j,2} \oplus \beta_{j,2} \oplus \cdots \oplus \alpha_{j,N} \oplus \beta_{j,N},
    \end{equation}
    where $N$ is the number of continuous columns and $\oplus$ denotes the concatenation operator. This mode specific technique ensures that the generated synthetic data $T_{\mathrm{syn}}$ follows the same distribution as the original data $T_{\mathrm{train}}$. The authors also verified that a classifier trained by $T_{\mathrm{syn}}$ achieved similar performance on $T_{\mathrm{test}}$ as a classifier learned on $T_{\mathrm{train}}$.

\begin{figure}[tb!]
\centering
  \includegraphics[scale = 0.195]{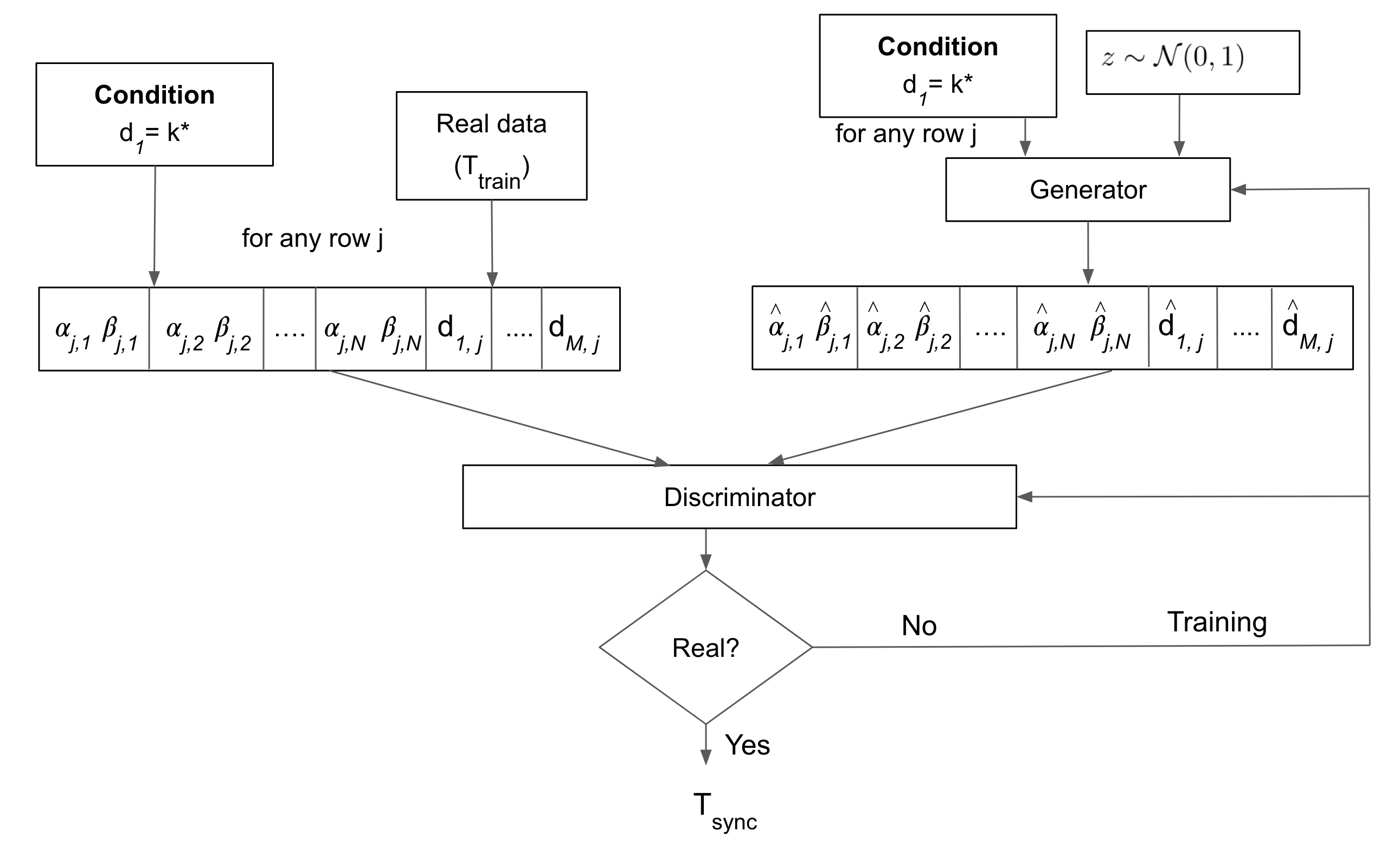}
  \caption{The CTGAN model: without loss of generality, it is assumed that the category $k^{*}$ in the first discrete column $d_{1}$ is selected as the condition to generate $T_{\mathrm{sync}}$.}
  \label{fig:ctgan modelt}
\end{figure}

For discrete columns, a conditional generation process is implemented to generate a new sample $\hat{r}$ based on a specific category of discrete columns. That is,
\begin{equation}
 \hat{r} \sim \mathbb{P}_{G}(\textrm{row} | D_{i}^{*} = k^{*}),   
\end{equation}
where $\mathbb{P}_{G}$ is the learnt distribution, $D_{i}^{*}$ is the $i$-th discrete column, and $k^{*}$ is one of the categories in that discrete column. The conditional generator must learn the conditional distribution well enough from the real data such that $\mathbb{P}_{G}(\textrm{row} | D_{i}^{*} = k^{*}) = \mathbb{P}(\textrm{row} | D_{i}^{*} = k^{*})$. This can be achieved by implementing a conditional vector, training by a sampling method, and minimizing the generator loss; see details in \cite{ctgan}. 

The overall CTGAN model is given in Figure \ref{fig:ctgan modelt}, where there are $N$ number of continuous columns and $M$ discrete columns. In our task, there are no categorical features. The only discrete column contains the labels indicating fire or no fire.

\subsection {Network architecture of CTGAN}
 Both the generator and the discriminator are fully connected neural networks capturing all possible correlations among all columns. There are two hidden layers in the generator and the discriminator, respectively. The rectified linear unit (ReLU) is used as the activation function in the generator while the leaky ReLU is used in the discriminator. There is another synthetic row representation layer in the generator after the two hidden layers. The scalar value in that layer $\alpha_{i}$ is generated by tanh function while the mode indicator $\beta_{i}$ is generated by the Gumbel-softmax continuous distribution, which  can be smoothly annealed into a categorical distribution.

\section{Results} \label{dis}
\begin{figure}[tb!]
\centering
  \includegraphics[scale=0.35]{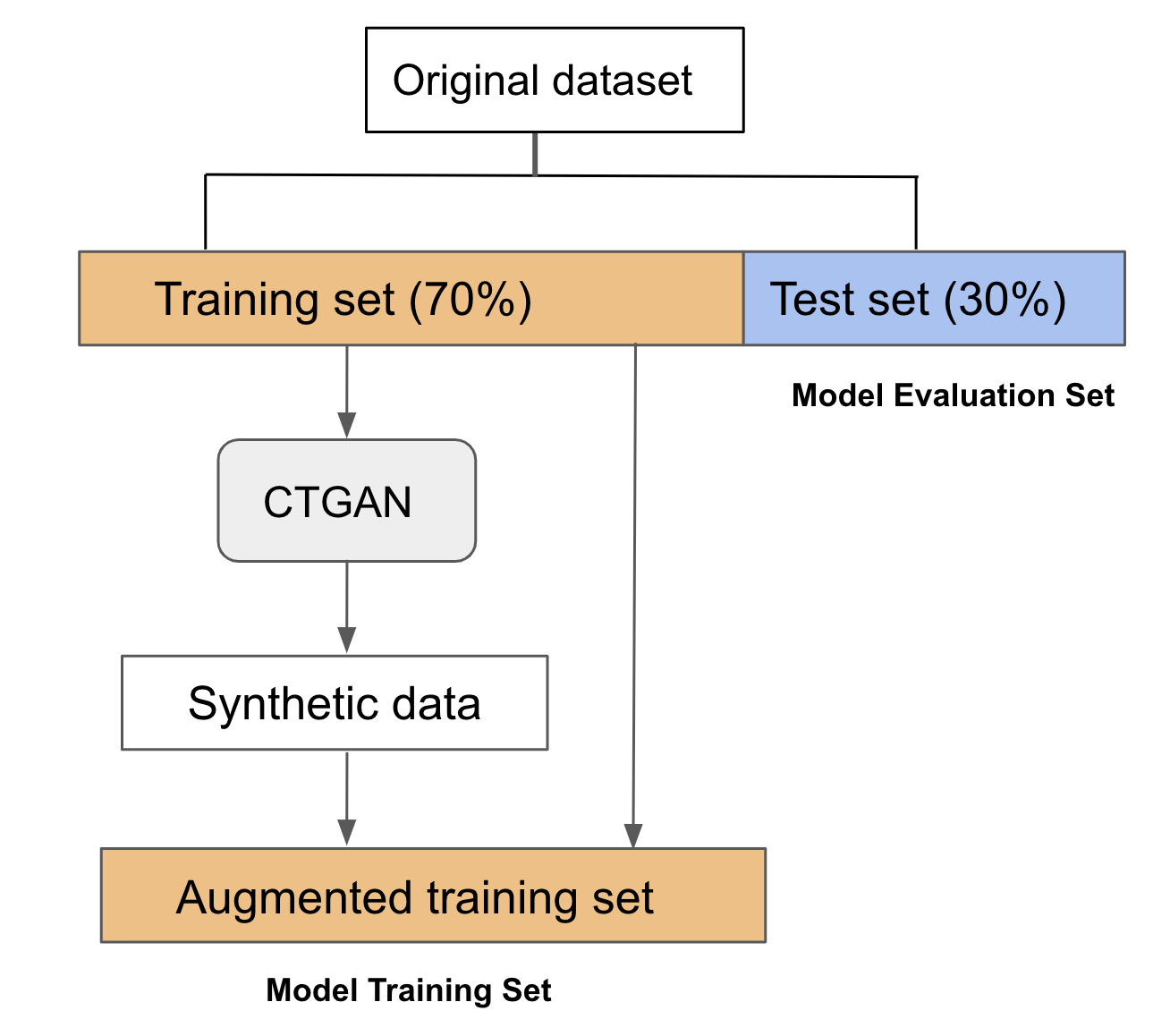}
  \caption{The segmentation of the training and test datasets.}
  \label{fig:parts of dataset}
\end{figure}

We first divide the constructed dataset (cf. section \ref{sec:DS formation}) into training and test set in a ratio of 70/30. The training set is used as the input of the CTGAN model. Then, the augmented dataset (i.e., generated synthetic data by the CTGAN and the original training set) is used to train all the ML models. Finally, we evaluate all models by using the original test set. The overall process is shown in Figure \ref{fig:parts of dataset}.

To show the merit of our proposed approach of CTGAN-based neural network (NN), we compare it with four other baseline models: decision tree (DT), random forest (RF), gradient boosting (GB), support vector machine (SVM), which are implemented by using Scikit-Learn 0.23.2. The NN is trained by using Tensorflow v2.0. 
The command ``sklearn.model-selection.GridSearchCV'' is used for tuning hyperparameters whose best values are given in Table~\ref{Tab:hypertuning}. 

 \begin{table} 
 \caption{The set of tuned hyperparameters. The bold values are the best ones with grid search hyperparameter tuning}
 \label{Tab:hypertuning}
 \centering
 \begin{tabular}{c|l}
 \hline
 Model & Hyperparameters \\  \hline 
 RF & Number of estimators (50,100,\textbf{200},600) \\ 
 & Maximum depth of each tree (5,10,\textbf{15},20) \\ 
 & Minimum samples leaf (1,\textbf{2},3,5) \\ 
 & Minimum samples split (1,3,5,10,\textbf{15}) \\   \hline
 SVM & Kernel (\textbf{RBF}, linear, polynomial) \\ 
 & Penalty parameters $C$ (0.1,\textbf{1},10)\\ 
 & Gamma $\gamma$ (0.1,1,\textbf{10})\\ \hline
 NN & Epochs (50,100,\textbf{200},300)\\
 & Batch size (2,8,16,\textbf{32},64)\\ 
 & Number of neurons in each layer (5,\textbf{10},20)\\
 & Learning rate (1e-1,1e-2,1e-3,\textbf{1e-4})\\
 & Optimizer (\textbf{Adam}, SGD)\\
 & Dropout rate (0,0.1,\textbf{0.2}) \\ \hline
\end{tabular}
\end{table}

\begin{figure}[tb!]
\center
  \includegraphics[scale=0.20]{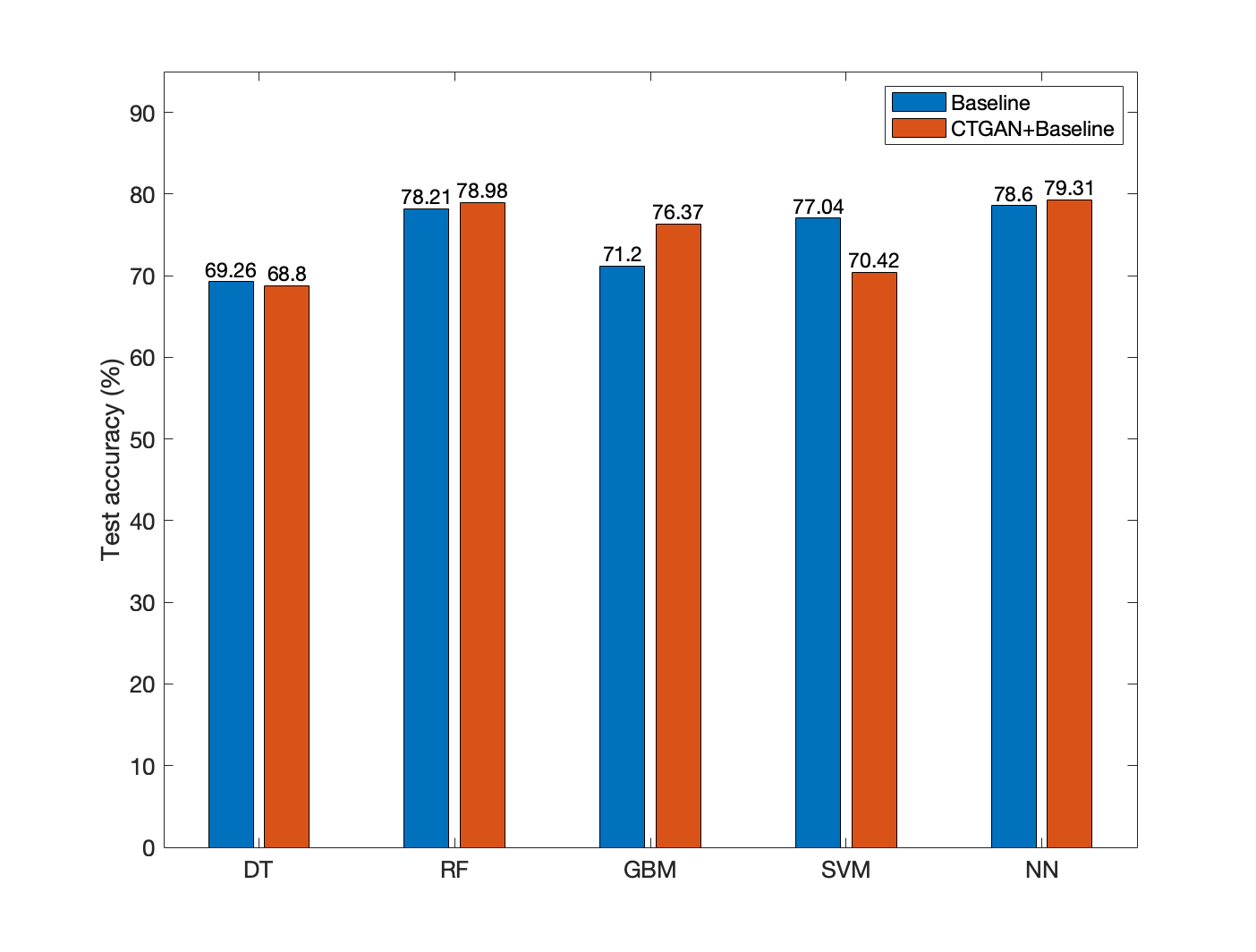}
  \caption{Performance comparison of baseline models and the proposed approach tested on the dataset that is described in section \ref{sec:DS formation}. }
  \label{fig:performace_our dataset}
\end{figure}

Figure \ref{fig:performace_our dataset} shows the results of  classification accuracy tested on our constructed dataset. 
It can be seen that the proposed CTGAN augmented NN achieves the best accuracy. In addition, we find that except for DT and SVM, the remaining three models with the CTGAN-augmented technique outperform their baseline counterparts.


To fully measure the performance of all models, we also leverage different metrics including the followings
\begin{align}
  \textrm{Precision} &= \frac{tp}{tp+fp} \\
  \textrm{Recall}  &= \frac{tp}{tp+fn} \\
  \textrm{F1 score}  &= 2\times \frac{\textrm{Precision} \times \textrm{Recall}}{\textrm{Precision}+\textrm{Recall}},  
\end{align}
where $tp$, $fp$, and $fn$ are the numbers of true positives, false positives, and false negatives, respectively.
Instead of using the vanilla version of these performance metrics, their counterparts of weighted average are used.   
For example, the weighted precision is defined as
\begin{equation}
        \bar{P}=\frac{S_{\mathrm{F}} \times P_{\mathrm{F}} + S_{\mathrm{NF}} \times P_{\mathrm{NF}}} {S_{\mathrm{F}}+S_{\mathrm{NF}}},
\end{equation} 
where $S_{\mathrm{F}}$ and $S_{\mathrm{NF}}$ are the number of samples in `Fire' and `No fire' class, respectively. $P_{\mathrm{F}}$ and $P_{\mathrm{NF}}$ denote the precision for the two classes. The weighted average of recall and F1 score are defined in a similar way. All results are summarized in Table~\ref{Tab:weightedmetric} that show the supremacy of CTGAN-based RF and NN.

\begin{table}[tb!]
\setlength{\tabcolsep}{5pt}
\caption{Performance comparison of weighted Precision, Recall and F1 score tested on the dataset in section \ref{sec:DS formation}}
 \label{Tab:weightedmetric}
 \centering
\begin{tabular}{*7c}
\cline{2-7}
 &  \multicolumn{3}{|c}{Baseline} & \multicolumn{3}{||c|}{Baseline+CTGAN}\\
\cline{1-7}

\multicolumn{1} {|c|} {Models} & {Precision}  &  {Recall} &   {F1} &  \multicolumn{1} {||c} {Precision} &  {Recall} &  \multicolumn{1} {c|} {F1} \\ \cline{1-7}

\multicolumn{1}{ |c|  }{DT}& 0.69 & 0.69 & 0.69 & \multicolumn{1} {||c} {0.69} & 0.69 & \multicolumn{1} {c|}  {0.69}\\  \cline{1-7} 
\multicolumn{1}{ |c|  }{RF}& 0.78 & 0.78 & 0.78 & \multicolumn{1} {||c} {0.80} & 0.79 & \multicolumn{1} {c|}  {0.79}\\ \cline{1-7} 
\multicolumn{1}{ |c|  }{GBM}& 0.71 & 0.71 & 0.71 & \multicolumn{1} {||c} {0.78} & 0.77 & \multicolumn{1} {c|}  {0.76}\\  \cline{1-7} 
\multicolumn{1}{ |c|  }{SVM}& 0.79 & 0.77 & 0.77 & \multicolumn{1} {||c} {0.73} & 0.70 & \multicolumn{1} {c|}  {0.70}\\  \cline{1-7} 
\multicolumn{1}{ |c|  }{NN}& 0.78 & 0.78 & 0.78 & \multicolumn{1} {||c} {0.79} & 0.78 & \multicolumn{1} {c|}  {0.79}\\ \cline{1-7} 
\end{tabular}
\end{table}

\begin{figure}[tb!]
\centering
  \includegraphics[scale=0.20]{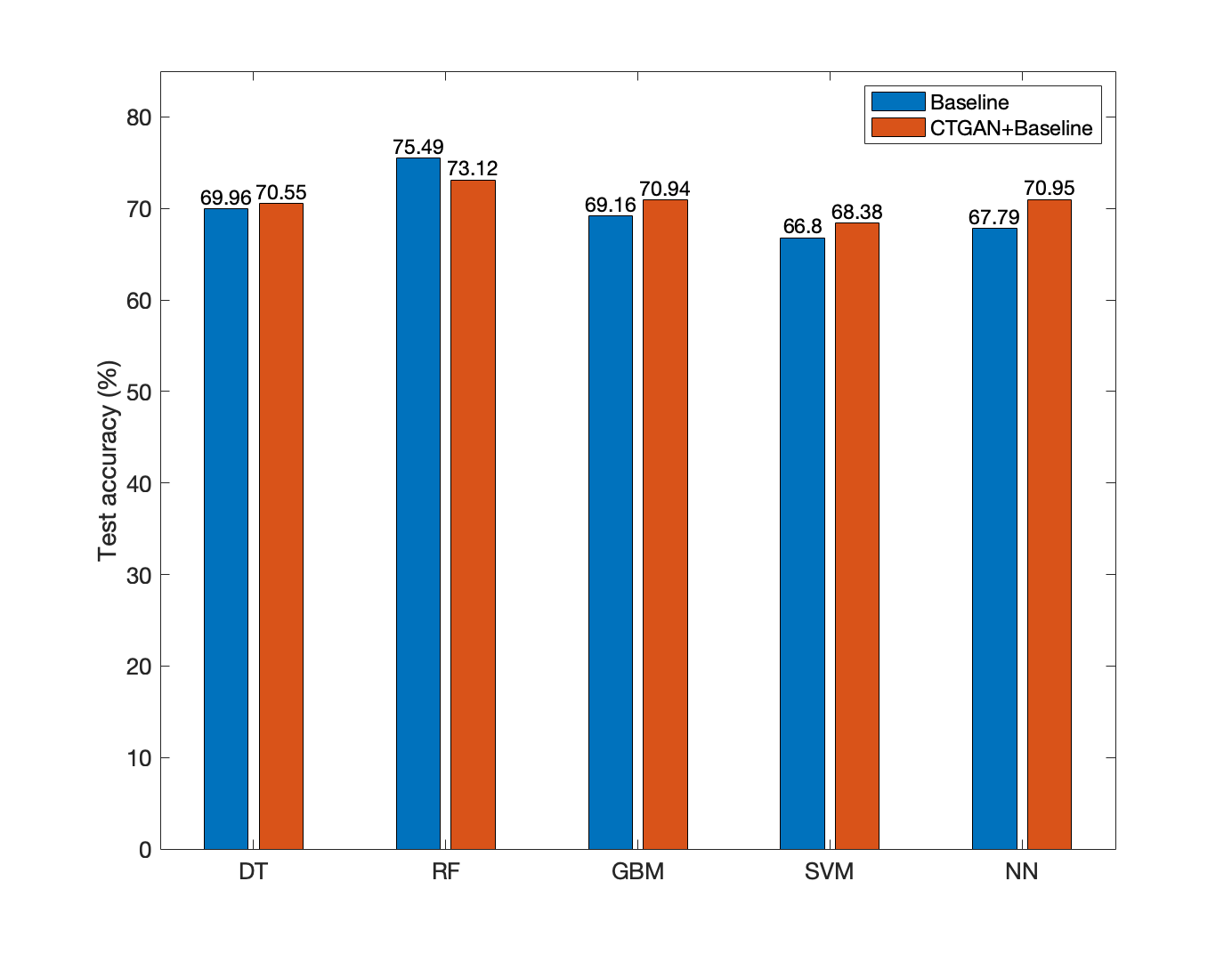}
  \caption{Performance comparison between baseline models and our proposed model on another dataset used by \cite{Sayad2019}. }\label{fig:performace_another dataset}
  \vspace{-0.5cm}
\end{figure}

\begin{table}[!tb]
\setlength{\tabcolsep}{5pt}
\caption{Performance comparison of weighted Precision, Recall and F1 score tested on the dataset in \cite{Sayad2019} }
 \label{Tab:weightedmetric_dataset2}
  \centering
\begin{tabular}{*7c}
\cline{2-7}
 &  \multicolumn{3}{|c}{Baseline} & \multicolumn{3}{||c|}{Baseline+CTGAN}\\
\cline{1-7}

\multicolumn{1} {|c|} {Models} & {Precision}  &  {Recall} &   {F1} &  \multicolumn{1} {||c} {Precision} &  {Recall} &  \multicolumn{1} {c|} {F1} \\ \cline{1-7}

\multicolumn{1}{ |c|  }{DT}& 0.68 & 0.70 & 0.68 & \multicolumn{1} {||c} {0.69} & 0.71 & \multicolumn{1} {c|}  {0.70}\\ \cline{1-7} 
\multicolumn{1}{ |c|  }{RF}& 0.72 & 0.73 & 0.71 & \multicolumn{1} {||c} {0.75} & 0.73 & \multicolumn{1} {c|}  {0.68}\\ \cline{1-7} 
\multicolumn{1}{ |c|  }{GBM}& 0.71 & 0.69 & 0.62 & \multicolumn{1} {||c} {0.70} & 0.71 & \multicolumn{1} {c|}  {0.69}\\  \cline{1-7} 
\multicolumn{1}{ |c|  }{SVM}& 0.74 & 0.67 & 0.55 & \multicolumn{1} {||c} {0.67} & 0.68 & \multicolumn{1} {c|}  {0.68}\\  \cline{1-7} 
\multicolumn{1}{ |c|  }{NN}& 0.71 & 0.68 & 0.58 & \multicolumn{1} {||c} {0.70} & 0.71 & \multicolumn{1} {c|}  {0.70}\\  \cline{1-7} 
\end{tabular}
\end{table}

Besides our newly constructed dataset, we also evaluate the performance of all those models in another dataset given in \cite{Sayad2019}. The dataset has two classes (`fire' and `no fire') with three features: burnt area, land surface temperature (LST), and NDVI. The result of test accuracy is presented in Figure \ref{fig:performace_another dataset}. Except for RF, all the models yield better results when trained with the augmented dataset. The values of the weighted precision, recall and F1 score are shown in Table~\ref{Tab:weightedmetric_dataset2}. 
Compared with the baseline cases, we can see that the proposed CTGAN-based approach helps improve the weighted recall and F1 score for all models 
except for the F1 score of RF.


\section{Conclusion} \label{conc}
In this paper, we propose a GAN-based approach to effectively predict wildfire risk by analyzing the underlying patterns among various features of fire ignition. First, we build a new dataset that consists of meteorological factors, vegetation indices, topology and distance from nearest powerline. Then, CTGAN is capitalized on to generate high-quality realistic data to facilitate different ML algorithms for better prediction performance. To show the effectiveness of our proposed approach, a comparative analysis is performed with five ML models, where the proposed CTGAN-based random forest and neural networks yield the best classification results, as tested on two different datasets.

Leveraging the proposed wildfire prediction approach, we can take several preventive and mitigation measures. For example, human activities should be restricted around susceptible fire locations since many large fires can be ignited by human ventures. Authorities can also carry out necessary vegetation management tasks like clearing bushes, performing prescribed burns, pruning risky branches of trees to prevent fires sparked by electric power infrastructure. During the preemptive shut down of power to prevent wildfire, utilities can also undertake the allocation of renewable energy fueled microgrids in order to meet the curtailed demand of their customers and at the same time, mitigating GHG emissions.
A few interesting research directions open up towards extending the proposed approach in this paper, we plan to conduct feature engineering that involves human activities, as well as wildland-urban interface on the projected rate of migration to develop a more effective and robust model for wildfire prediction.


\nocite{*}
\bibliographystyle{IEEEtran}
\bibliography{LSrefs,IEEEabrv}

\end{document}